\documentclass[letter, 10 pt, conference]{ieeeconf}  % Comment this line out if you need a4paper

\IEEEoverridecommandlockouts                              % This command is only needed if
\usepackage{amsmath}
\usepackage{amssymb}
\usepackage{algorithmicx, algpseudocode}
\usepackage{mathtools}
\usepackage{graphicx}
\usepackage{float}
\usepackage{array}
\usepackage{tikz}
\usepackage{listings}
\usepackage{hyperref}
\usepackage{graphicx}
\usepackage{cite}
\usepackage{dsfont}
\usepackage{mathtools,etoolbox}
\usepackage[ruled]{algorithm2e}
\usepackage[none]{hyphenat}
\usepackage[utf8]{inputenc}
\usepackage[english]{babel}

\usepackage{amsthm}
\usepackage{xfrac}
\usepackage{nicefrac}
\usepackage{booktabs}
\usepackage{flushend}
\usepackage{makecell}
\usepackage{tikz-cd}
\usepackage{multirow}

\DeclareMathAlphabet{\mathpzc}{OT1}{pzc}{m}{it}
\newcommand{\etal}{\textit{et al.}}

\DeclareMathOperator*{\argmin}{argmin} % no space, limits underneath in displays
\DeclareMathOperator*{\argmax}{argmax} % thin space, limits underneath in displays

\hypersetup{colorlinks=true,urlcolor=blue,citecolor=red}

\newcommand{\norm}[1]{\left\lVert#1\right\rVert} % Command for auto adjust norm!

\title{\LARGE \bf 0-MMS: Zero-Shot Multi-Motion Segmentation With A\\ Monocular Event Camera}
\author{Chethan M. Parameshwara, Nitin J. Sanket, Chahat Deep Singh, Cornelia Ferm{\"u}ller, Yiannis Aloimonos % <-this % stops a space
\thanks{This work was supported in parts by the Office of Naval Research and the the National Science Foundation  under Grants N00014-17-1-2622 and BCS 1824198 respectively and by Samsung Electronics. \textit{Corresponding Author: Chethan M. Parameshwara}.}
\thanks{All authors are associated with the Perception and Robotics Group, University of Maryland, College Park. Emails: \{{\tt\footnotesize cmparam9, nitin, chahat, fer, yiannis}\} {\tt \footnotesize @umiacs.umd.edu}}} % <-this % stops a space

\begin{document}

\makeatletter
\g@addto@macro\@maketitle{
\begin{figure}[H]
  \setlength{\linewidth}{\textwidth}
  \setlength{\hsize}{\textwidth}
    \centering
    \includegraphics[width=\textwidth]{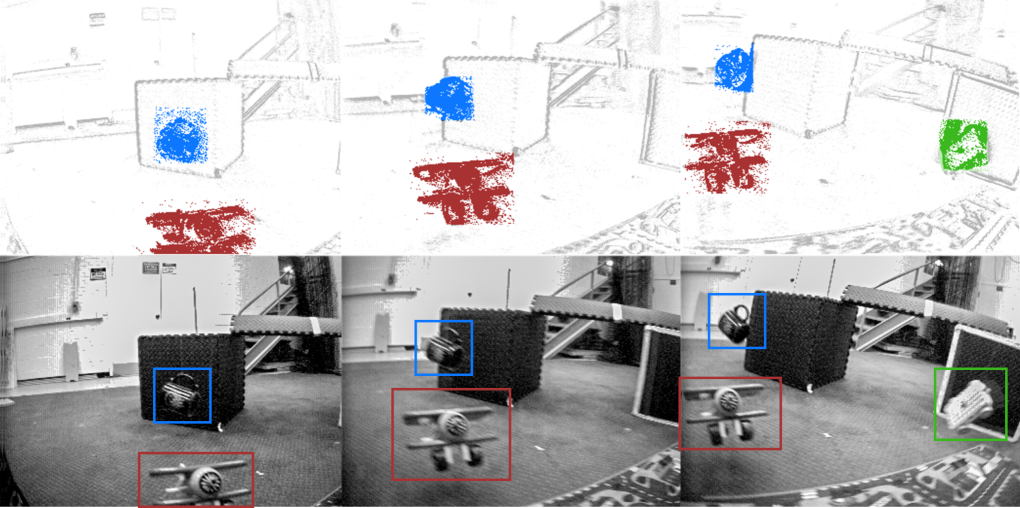}
    \caption{Multi-Motion Segmentation with a monocular event camera on an EV-IMO dataset sequence. Top Row: The event frames are color-coded by cluster membership. The
    corresponding grayscale frames are shown in the bottom row. Bounding boxes on the images are color coded with respect to the objects for reference. \textit{Note that grayscale images are not used for computation and are provided for visualization purposes only}. \textbf{All the images in this paper are best viewed in color.}}
    \vspace{-20pt}
    \label{fig:MOMSWithEventsBanner}
    \end{figure}
}
\maketitle
\thispagestyle{empty}
\pagestyle{empty}
% \thispagestyle{plain}
% \pagestyle{plain}

%% Setting figure number to 2. Adding figure to \maketitle counts as two Fig. for some weird reason!!!
% \setcounter{figure}{1}

%%%%%%%%%%%%%%%%%%%%%%%%%%%%%%%%%%%%%%%%%%%%%%%%%%%%%%%%%%%%%%%%%%%%%%%%%%%%%%%%%%%%%
\begin{abstract}
%1. Why we do it?
Segmentation of moving objects in dynamic scenes is a key process in scene understanding for navigation tasks.
Classical cameras suffer from motion blur in such scenarios rendering them effete. On the contrary, event cameras, because of their high temporal resolution and lack of motion blur, are tailor-made for this problem.
% Without prior knowledge of the object structure and motion, the problem is very challenging due to  the plethora of motion parameters to be estimated while being agnostic to motion blur and occlusions. Event cameras, because of their high temporal resolution, and lack of motion blur, are tailor made this problem.
We present an approach for monocular multi-motion segmentation, which combines bottom-up feature tracking and top-down motion compensation into a unified pipeline, which is the first of its kind to our knowledge. Using the events within a time-interval, our method segments the scene into multiple motions by splitting and merging. We further speed up our method by using the concept of motion propagation and cluster keyslices.

The approach was successfully evaluated on both challenging real-world and synthetic scenarios from the EV-IMO, EED, and MOD datasets and outperformed the state-of-the-art detection rate by 12\%, achieving a new state-of-the-art average detection rate of 81.06\%, 94.2\% and 82.35\% on the aforementioned datasets. To enable further research and systematic evaluation of multi-motion segmentation, we present and open-source a new dataset/benchmark called MOD++, which includes challenging sequences and extensive data stratification in-terms of camera and object motion, velocity magnitudes, direction, and rotational speeds.

%We achieve 73.21\% detection rate on MOD++ which is 2 to 3$\times$ higher than the state-of-the-art methods.
% \textbf{We also present MOD++}.

% Top down model fitting with bottom-up point features

%Multi-object motion segmentation in a dynamic and unstructured environment is a key step for outlier removal and scene understanding. The difficulty of this problem is exacerbated by the amount of motion blur and lack of prior knowledge of the object structure and motion.
% A class of sensors that are particularly suitable for such scenarios are event cameras.
%2. What we do
%A class of sensors that are particularly suitable for such scenarios are event cameras. We propose a solution to the aforementioned problem without prior knowledge of moving objects and scene. Our method fragments the scene into multiple motion segments which are then iteratively merged by minimizing temporal gradients and maximizing contrast.

% We solve multi motion segmentation problem without prior knowledge of moving objects and scenes using the novel neuromorphic event sensors. Our method segments the scene into multiple motion iteratively fitting and merging models by maximizing contrast and minimizing temporal gradients.
%3. What do we achieve?
%We successfully evaluate our approach on both synthetic and challenging real-world scenarios contained in EED, MOD, EV-IMO datasets. We obtain a remarkable accuracy of \% x. To our knowledge, this is the first work on multi-object motion segmentation without prior knowledge on the number or structure of objects using event cameras.

\end{abstract}

\setcounter{figure}{1}

% \textbf{\textit{\small{Keywords -- Motion and Tracking, Vision for Robotics, Deep Learning: Applications, Methodology, and Theory, Scene Understanding, Segmentation, Grouping and Shape}}}
%%%%%%%%%%%%%%%%%%%%%%%%%%%%%%%%%%%%%%%%%%%%%%%%%%%%%%%%%%%%%%%%%%%%%%%%%%%%%%%%%%%%%
% SUPPLEMENTARY MATERIAL
% The accompanying video is available at

\section*{Supplementary Material}
The accompanying video
and dataset are available at \url{prg.cs.umd.edu/0-MMS.html}.

%%%%%%%%%%%%%%%%%%%%%%%%%%%%%%%%%%%%%%%%
% TODO
% 06 Oct 2020
% \begin{enumerate}
%     \item Add cluster splitting
%     \item Add keyslices for speedup
%     \item Add object breaking into multiple pieces into MOD++ and track each part (Borrow ideas and some datasets from http://rpg.ifi.uzh.ch/E2VID.html)
%     \item Compare optical flow EVFlownet, frames using e2vid trained on solomo with our classical method
% \end{enumerate}
% 17 Oct 2020
% \begin{enumerate}
%     \item Math/Algorithm Explanation
%     \begin{enumerate}
%         \item Tracklet definition
%         \item Event motion compensation
%         \item Event cloud definition
%         \item Split and Merge Algorithm (1)
%         \item Scene/cluster blur check (Eq.)
%         \item Cluster Propogation Algorithm (2)
%     \end{enumerate}
% \end{enumerate}
% 20 Oct 2020
% \begin{enumerate}
% % \item Banner Image: Exploding sequence and Collision Sequence
% % \item Stepwise outputs
% % \item Num of objects detected vs real for exploding and collision sequence
% % \item MOD++ evaluation table
% % \item MOD++ results image for different sequences
% \end{enumerate}

% \begin{enumerate}
%     \item Refine Intro
%     \item Cut Related Work and Number of Refs to 25
%     \item Write about MOD++
%     \item EVIMO - Stratification
%     \item Ask Stoffregen for results on MOD++
%     \item Compute speed on GPU
%     \itemm Remove Lambda
%     \item Performance change with number of events (small vs large integration time)
% \end{enumerate}

\section{Introduction}

% \begin{figure}
%     \includegraphics[width=\columnwidth]{0-MMS/images/Banner.png}
% % \vspace{-1.3\baselineskip
%   \caption{Multi-Object Motion Segmentation with a monocular event camera. (a) Grayscale image (\textit{for illustration only, not used by the approach}) (b) Projection of a spatio-temporal event cloud (the input to the method is the event cloud). (c) Output event clusters, gray indicates background, red, green and blue boxes and colors indicate the different segmented moving objects. \textbf{All the images in this paper are best viewed in color.}}
% \label{fig:2}
% \end{figure}

Navigation is a fundamental competence of life with visual motion estimation as its beating heart\cite{fermuller1993navigational, sanket2018gapflyt}. Even though motion estimation has seen a tremendous advancement in the last few decades, dynamic object motion is usually addressed by outlier rejection schemes as a part of the mature structure from motion and SLAM pipelines \cite{liang2019salientdso}. Though, this can provide an initial dynamic object segmentation, further processing for each segment relies on some prior information (commonly appearance/structure/recognition).
% which are often targeted towards static scenes or when specific moving objects are known in advance.

To exacerbate the scenario further, classical imaging cameras often fail in dynamic scenarios (moving objects) due to motion blur and low light scenarios. To this end, drawing inspiration from nature, neuromorphic engineers developed a sensor called \textit{Dynamic Vision Sensor} (DVS) \cite{lichtsteiner2008128} which records the asynchronous temporal changes in the scene in the form of a stream of events, rather than the conventional image frames. This gives an unparalleled advantage in-terms of temporal resolution, low latency, and low band-width signals. Such event data is tailor-made for motion segmentation because of the disparity in event density at object boundaries.

In this paper, we present a method to detect moving objects by inferring their motion using a monocular event camera; we call this \textit{multi-motion segmentation}. We formally define the problem statement and our contributions next.

\subsection{Problem Formulation and Contributions}
We address the following question: \textit{How do you cluster the scene into background and Independently Moving Objects (IMOs) based on motion using data from a moving monocular event camera?} % without a prior of object shape, size, structure and number of objects?

Given an event volume $\mathcal{E}$, we find and cluster the events based on 2D motion. We over-segment the scene with the help of feature tracks and then merge clusters based on the motion models and a contrast score. Each cluster is represented by a four parameter motion model (denoting the similarity transformation/warp) $\Theta = \{\Theta_x, \Theta_y, \Theta_z, \Theta_\theta \}$ which represents the 2D translation ($\Theta_x, \Theta_y$), divergence and in-plane rotation, respectively \cite{iROSBetterFlow}. To speed up computation, we propagate these motion models until a cluster keyslice is invoked. A summary of our contributions is given below (Sample outputs are shown in Fig. \ref{fig:MOMSWithEventsBanner}):

\begin{itemize}
    \item A novel cluster splitting and merging approach for monocular event-based multi-motion segmentation without prior knowledge of scene geometry (zero-shot) and a number of objects.
    \item New open-source multi-motion segmentation dataset and benchmark MOD++ including extensive motion stratification and including challenging collision/exploding sequences.
    \item Speeding up computation using motion propagation and introduction of cluster keyslices.
\end{itemize}

% We make the following assumptions:
% \begin{enumerate}
%     \item The magnitude of IMO motion is slightly larger than the ego-motion.
%     \item  Ego-motion is not dominated by rotation.
% \end{enumerate}

% \section{Related Work}
\subsection{Related Work}
%Compensation approaches(Top-down approaches)}
There has been extensive progress in the field of event-based motion segmentation in the past decade for different scenarios at variable scene complexity. Earlier works focused on the case of a static camera, where the events are generated by the moving objects, and a simple clustering scheme can provide motion segmentation \cite{litzenberger2006,linares2015usb3,mishra2017saccade,barranco2018real}. The next mark up in complexity is the case of a moving camera, where event alignment is computed for the whole scene \cite{gallego2017accurate,zhu2017event,iROSBetterFlow} (also called sharpness or contrast measure \cite{gallego2019focus,Stoffregen_2019_CVPR}) and the parts of the scene which are misaligned give the motion segmentation of IMOs \cite{iROSBetterFlow} using a simple thresholding algorithm. The results were further improved by \cite{stoffregen2019event}, who used an Expectation-Maximization scheme to obtain better segmentation. Our work is closely related to \cite{stoffregen2019event} with the same underlying philosophy: using motion compensation for clustering but adds robustness in long term segmentation using feature tracking and cluster splitting and merging. We also introduce motion propagation and the concept of cluster keyslices to speed-up the entire procedure.

Finally, a few approaches used machine learning. \cite{barranco2015bio} learned object contours and border-ownership information via a structured random forest, which they demonstrated for segmentation. \cite{sanket2019evdodgenet, mitrokhin2019ev} demonstrated a combination of supervised and unsupervised CNN learning using deblurring/event-alignment in the cost function, and \cite{mitrokhin2020learning} designed a graph convolutional neural network for supervised motion segmentation, that uses as input event volumes over extended time periods.

\section{Preliminaries}

\subsection{Data From An Event Camera}

A traditional camera records frames at a fixed frame rate by integrating the number of photons for the chosen shutter time for all pixels \textit{synchronously}. In contrast, an event camera only records the polarity of logarithmic brightness changes \textit{asynchronously} at each pixel. If the brightness at time $t$ of a pixel at location $\mathbf{x}$ is given by $I_{t,\mathbf{x}}$ an event is triggered when $\Vert \log\left( I_{t+\delta t,\mathbf{x}}\right) - \log\left( I_{t,\mathbf{x}}\right) \Vert_1 \ge \tau $. Here, $\delta t$ is a small time increment and $\tau$ is a threshold which will determine the trigger of an event ($\tau$ is set at the driver level as a combination of multiple parameters). Each event outputs the following data:  $\mathbf{e} = \left\{\mathbf{x}, t, p \right\}$, where $p = \pm 1$ denotes the sign of the brightness change. We'll denote events in a spatio-temporal window as $\mathcal{E}_t = \{e_i\}^N_{i=1}$ ($N$ is the number of events) and we'll refer to  $\mathcal{E}$ as event slice/stream/cloud/volume.

\subsection{Model Fitting For Contrast Maximization}

Processing event cloud is generally computationally very expensive and to speed up the processing we use a projection function. The projection of $\mathcal{E}$ leads to a ``blurry'' image, and a number of methods for measuring this blurriness to achieve event-cloud alignment (also called contrast maximization or motion compensation or deblurring) have been developed \cite{Gallego2018AUC, gallego2019focus,Stoffregen_2019_CVPR}. The output of the alignment is an event frame denoted as $\mathpzc{E}_t$. In particular, we utilize the method proposed in \cite{iROSBetterFlow} to estimate model parameters $\Theta$ to maximize the alignment $\mathcal{E}$ by minimizing temporal gradients $\nabla \mathcal{T}$. Here $\mathcal{T}\left( \mathcal{E}\right) =\mathbb{E}\left(t_{\mathbf{x}} - t_0 \right)$, $\mathbb{E}$ is the expectation/averaging operator, $t_{\mathbf{x}}$ denotes the time value at location $\mathbf{x}$, and $t_0$ is the initial time of the temporal window. Formally, we solve the following optimization problem: $\argmin_{\Theta} \Vert \nabla \mathcal{T} \Vert_2 $, where $\nabla$ denotes the spatial gradient operator. Note that the projection function denoted by $\mathcal{W}$ is also called a warping function and refers to the creation of $\mathpzc{E}_t$ using parameters $\Theta$.

% This estimates the warp parameters $\Theta$ which maximizes the contrast or sharpness, thereby minimizing the temporal gradients.
% \begin{equation}
% \mathcal{T}\left( \mathcal{E}\right) = \left(\sum_{t  = t_0}^{t_0 + \delta t} \mathbbm{1}\left(\mathcal{E}\left( \mathbf{x}, t, p=\pm 1\right)\right) \right)^{-1} \mathbb{E}\left( t - t_0\right)
% \label{eq:T}
% \end{equation}

% \begin{equation}
% \mathcal{T}\left( \mathcal{E}\right) =\mathbb{E}\left(t_{\mathbf{x}} - t_0 \right) %\mathbbm{1}\left(\mathcal{E}\left( \mathbf{x}, t_{\mathbf{x}}, p=\pm 1\right)\right)
% \label{eq:T}
% \end{equation}

\subsection{Tracklets Using Point Tracker}
% Once the event cloud $\mathcal{E}$ is warped using $\Theta$ and then projected onto the 2 dimensional image plane, we obtain a motion compensated image $\mathpzc{E}$ which is sharp on the background boundaries and ``blurry'' on the object boundaries (denoting the inconsistency in the motion model between the background and IMOs). Grouping these inconsistencies is not trivial and requires local motion information. Hence, we employ a sparse feature detection and tracking approach to gather local information and group residual motion. Over the past few years, robust feature extraction and tracking approaches for event data have been proposed.  We found deep learning-based approaches to be more robust and generalizable over a wide range of scenarios without data fine tuning. Hence, we utilize the approach of DeTone \etal~ \cite{detone18superpoint} (previously used on gray-level images) for extracting sparse features $\mathpzc{F}$ from $\mathpzc{E}$.   % In the next step, $\mathcal{F}$ is used for grouping into multiple feature clusters  $\{\mathcal{P}_j\}^K_{j=1}$ in an iterative model fitting and merging step.\\
We rely on obtaining tracklets (feature tracks across multiple event frames) as an input to multi-motion segmentation. Over the past few years, robust feature extraction and tracking approaches for event data have been proposed \cite{seok2020robust, gehrig2020eklt, dardelet2018event, zhu2017event, alzugaray2019asynchronous}, however most of the robust methods are relatively slow or not open-source or use conventional intensity images. Hence, we adapt  SuperPoint \cite{detone18superpoint} (previously used on grayscale images) for extracting tracklets $\mathbb{T}_t$ from a set  of consecutive $N$ event frames $\{\mathpzc{E}_{t+\Delta t\times i} \vert i\in[0,N-1]\}$. We found the SuperPoint tracker to be robust and generalizable over a wide range of scenarios without any fine tuning. The SuperPoint tracker runs in the backend (we call this Tracker backend) on a First-In First-Out (FIFO) buffer of consecutive $N$ event slices.

% We use $N$ consecutive event frames $\{\mathpzc{E}_{t+\Delta t\times i} \vert i\in[0,N-1]\}$ to compute tracklets ($\mathbb{F}_t$).

% We found tdeep learning-based approaches to be more robust and generalizable over a wide range of scenarios without data fine tuning. Hence, we utilize the approach of DeTone \etal~ \cite{detone18superpoint} (previously used on gray-level images) for extracting sparse features $\mathpzc{F}$ from $\mathpzc{E}$.

\section{Proposed Approach}

\subsection{Overview}

\begin{figure*}
\begin{center}
    \includegraphics[width=\textwidth]{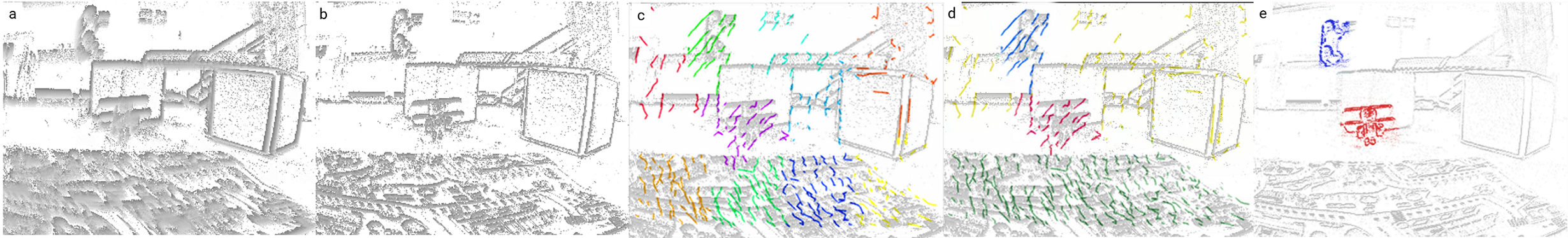}
\end{center}
 % \vspace{-1.3\baselineskip}
   \caption{\small{ Overview of the proposed pipeline on a sequence from EV-IMO dataset (a) Projection of the raw event cloud $\mathcal{E}_t$ without motion compensation, (b) Projection of event cloud after global motion compensation ($\mathpzc{E}_t$), (c) Sparse tracklets $\mathbb{F}_t$ extracted on compensated event cloud, (d) Merged feature clusters based on contrast and distance metrics ($\{ \mathbb{C}_t\}$),  (e) Output of the pipeline is the cluster of events. The cluster membership is color coded where gray color indicating background cluster.}}
\label{fig:pipeline}
\end{figure*}

 The proposed solution comprises of two steps: 1. \textit{Spilt and Merge} summarized in Algorithm \ref{alg:splitandmerge} and illustrated in Fig. \ref{fig:pipeline}. 2. \textit{Motion Propagation and Cluster keyslices} summarized in Algorithm \ref{alg:propagation}.

\begin{algorithm}[t!]
\caption{Splitting and Merging}
\label{alg:splitandmerge}
\SetAlgoLined
\KwData{Tracklets $\mathbb{T}_t$, Event Stream $\mathcal{E}_t$, Num. oversegments $K$}
\KwResult{Clusters $\{\mathbb{C}_t\}$, Cluster Models

$\{\Theta_t\}$, Segmentation Masks $\{\mathcal{S}\}$

}
\texttt{Splitting}\;
$\{\mathbb{\widetilde{C}}_t\}$ = $k$-Means($\mathbb{T}_t$, $K$)\;
\texttt{Merging}\;
$\{\widetilde{\Theta}_t\}$ = ClusterModelFit($\mathcal{E}_t$, $\{\mathbb{\widetilde{C}}_t\}$)\;
 \While{Stopping Criterion and All Clusters Visited}{
 \If{$\mathcal{C}_{k,j} \mathcal{D}_{k,j}^{-1} > \zeta$ (\Comment{Merging Criterion})}{
$\{\mathbb{\widetilde{C}}_t\}, \{\widetilde{\Theta}_t\}$ = MergeClusters($\mathcal{E}, \{\mathbb{\widetilde{C}}_t\}, \{\widetilde{\Theta}_t\}$) \hfill \Comment{Updated Clusters and Motion Models}\;}}
$\{\mathbb{C}_t\}$ = $\{\mathbb{\widetilde{C}}_t\}$ \hfill \Comment{Final Clusters}\;
$\{\Theta_t\}$ = $\{\widetilde{\Theta}_t\}$ \hfill  \Comment{Final Models}\;
$\mathcal{S}$ = ConvexHull($\{\mathbb{C}_t\}$) \Comment{Final Dense Segmentation}\;
\end{algorithm}

\begin{algorithm}[t!]
\caption{Motion Propagation And Cluster Keyslices}
\label{alg:propagation}
\SetAlgoLined
\KwData{Tracklets $\mathbb{T}_t$, Event Stream $\mathcal{E}_t$, Clusters $\{\mathbb{C}_{t-1}\}$, Cluster Models $\{\Theta_{t-1}\}$}
\KwResult{Clusters $\{\mathbb{C}_{t}\}$, Cluster Models $\{\Theta_{t}\}$}
\ForEach{Cluster $i$}{
$\mathpzc{E}_t^i = \mathcal{W}(\mathcal{E}_t^i, \mathbb{C}_{t-1}^i, \Theta_{t-1}^i)$ \hfill \Comment{Cluster Event frame}\;
}
\eIf{$\mathbb{E}(\mathcal{C}(\mathpzc{E}_t^i)\,\,\forall i) > \tau $ \Comment{Scene Contrast Measure}\;}{
\ForEach{Cluster $i$}{
\eIf{$\mathcal{C}(\mathpzc{E}_t^i) > \chi $}{
Keep Propagation\; \Comment{no new cluster keyslice required}\;
}{
Split and Merge on current cluster (Algorithm \ref{alg:splitandmerge})\; \Comment{New cluster keyslice\;}
}}}
{
Split and Merge on entire scene (Algorithm \ref{alg:splitandmerge})\;
\Comment{New cluster keyslice for all clusters (scene)\;}
}
\end{algorithm}

\subsection{Split And Merge}
% The split and merge step consists of five steps, which are described next and summarized in Algorithm \ref{alg:main} and Fig. \ref{fig:pipeline}{\color{red}a}.

\noindent \textbf{Splitting}:  Here, the tracklets from the backend are clustered into $k$ clusters ($k >> $ Num. of objects) using $k$-Means clustering for its simplicity and speed. If a prior on the number of objects or a bound is known, it can be trivially incorporated to choose $k$. We denote these clusters as $\mathbb{C}_t$.\\

% The first step is to group features based on motion.

% We employ a simple K-means clustering for its simplicity and speed to cluster the feature tracks (commonly called tracklets) based on the length of the tracks and spatial location.

\noindent \textbf{Merging}: Since the splitting method oversegments the scene, we need to merge the clusters to obtain motion segmentation for Independently Moving Objects (IMOs) and the background. The cluster merging is based on a similarity measure that depends on the contrast match (warping a cluster with the model from another cluster and measuring the contrast) and distance between centroids of the clusters. We define contrast and distance functions $\mathcal{C}_{k, j}$ and $\mathcal{D}_{k, j}$ respectively as follows: $\mathcal{C}_{k, j} = \mathbb{E}\left(\Vert\text{Var}\left(\mathpzc{E}(\delta \mathcal{E}_j | \Theta_k) \right)\Vert_1\right)$ and $\mathcal{D}_{k, j} = \Vert\mathbf{C}_{k} - \mathbf{C}_{j} \Vert_2 $ where $k, j$ are the cluster numbers, $\delta \mathcal{E}_j$ represents the event volume for cluster $j$ and $\mathbf{C}_k$ denotes the centroid of cluster $k$. This formally entails solving the following optimization problem: $\argmax_{j}\,\, \mathcal{C}_{k,j} {\mathcal{D}}^{-1}_{k,j}$ which simultaneously maximizes the contrast and minimizes the distance. This step is iteratively performed per cluster (where merging happens with every neighboring cluster using breath first search) until a stopping criterion has been reached. The entire process is repeated until all the clusters have been visited. The stopping criterion is explained next.

After each merging operation, we compute the motion model $\Theta_{k,j}$ of the merged clusters by minimizing the temporal gradients $\nabla \mathcal{T}$. Intuitively, when two clusters are merged, the combined motion model captures the average motion of the two clusters thereby slightly increasing the average temporal gradients. Further, whenever a moving object cluster is merged with the background or different IMO cluster the average temporal gradient increases drastically. Hence, a difference in temporal gradient at every step $i$ ($  \mathbb{E}\left(\Vert \nabla \mathcal{T}_i\Vert_2\right)$) with respect to the initial step ($\mathbb{E}\left(\Vert \nabla \mathcal{T}_0\Vert_2\right)$) is computed. If at any step the difference in temporal gradient is large, we terminate the current merge and continue to the next iteration until all the clusters have been visited. This is mathematically described by $ \Vert \mathbb{E}\left(\Vert \nabla \mathcal{T}_i\Vert_2\right) - \mathbb{E}\left(\Vert \nabla \mathcal{T}_0\Vert_2\right) \Vert_1 \ge \lambda $, where $\lambda$ is some constant threshold.

After split and merge has been performed, we obtain the final clusters $\{\mathbb{C}_t\}$ ($\{ \}$ indicates a set of clusters and each cluster can be indexed with a superscript, i.e., $\mathbb{C}_t^i$ for $i^{\text{th}}$ cluster) and motion models per cluster $\Theta_t^i$ where $i$ indexes the cluster number. Optionally, we obtain the dense segmentation by taking the convex hull of the sparse feature points in each cluster, which is denoted as $\mathcal{S}$. Refer to Algorithm \ref{alg:splitandmerge} for a summary of split and merge methods. % The entire procedure is initialized with this procedure.

\subsection{Motion Propagation}
Optimizing the parameters $\Theta$ at every time step to obtain $\mathpzc{E}_t$ from $\mathcal{E}_t$ is computationally exorbitant. Inspired by classical tracking pipelines, we propagate motion models from previous to current time slice assuming linear event trajectories. The propagation is achieved using tracklets, and we validate the propagation quality using the following contrast function $\mathcal{C}_{t} = \mathbb{E}\left(\Vert\text{Var}\left(\mathpzc{E}(\delta \mathcal{E}_{t} | \Theta_{t-1}) \right)\Vert_1\right)$. Here, $\mathcal{C}_{t}$ measures the deviation from the current optimal motion model with respect to the motion model of the previous time slice.

% TODO: Talk about how much speedup you obtain

\subsection{Speeding-Up Computation Using Cluster Keyslices}
% Visual Odometry
Classical Visual Odometry pipeline utilizes the concept of keyframes to speed-up computation based on certain conditions, this avoids performing bundle adjustment on every frame whilst maintaining good accuracy. We employ a similar strategy of keyframes for event slices, which we call a \textit{keyslice}, for re-clustering based on contrast function $\mathcal{C}_{t}$. We apply the contrast function at different levels starting from the entire scene to each clusters separately. Depending on the following two measures the split and merge procedures are performed either per cluster or on the entire scene: 1. Cluster contrast score, 2. Scene contrast score. The cluster contrast score is defined by $\mathcal{C}_{t}$ when applied to a single cluster during motion propagation and the average of all cluster scores is called the scene contrast score. Refer to Algorithm \ref{alg:propagation} for a summary of motion propagation and cluster keyslicing.

% TODO: Speedup achieved

% Based on the variation of contrast, we initiate the split and merge step again on cluster keyslices. Otherwise,

% keyslice.

% Constrast score for scene.

% \subsection{Initialization}

% TODO: Write about speedup

\begin{figure}[t!]
\begin{center}
    \includegraphics[width=0.9\columnwidth]{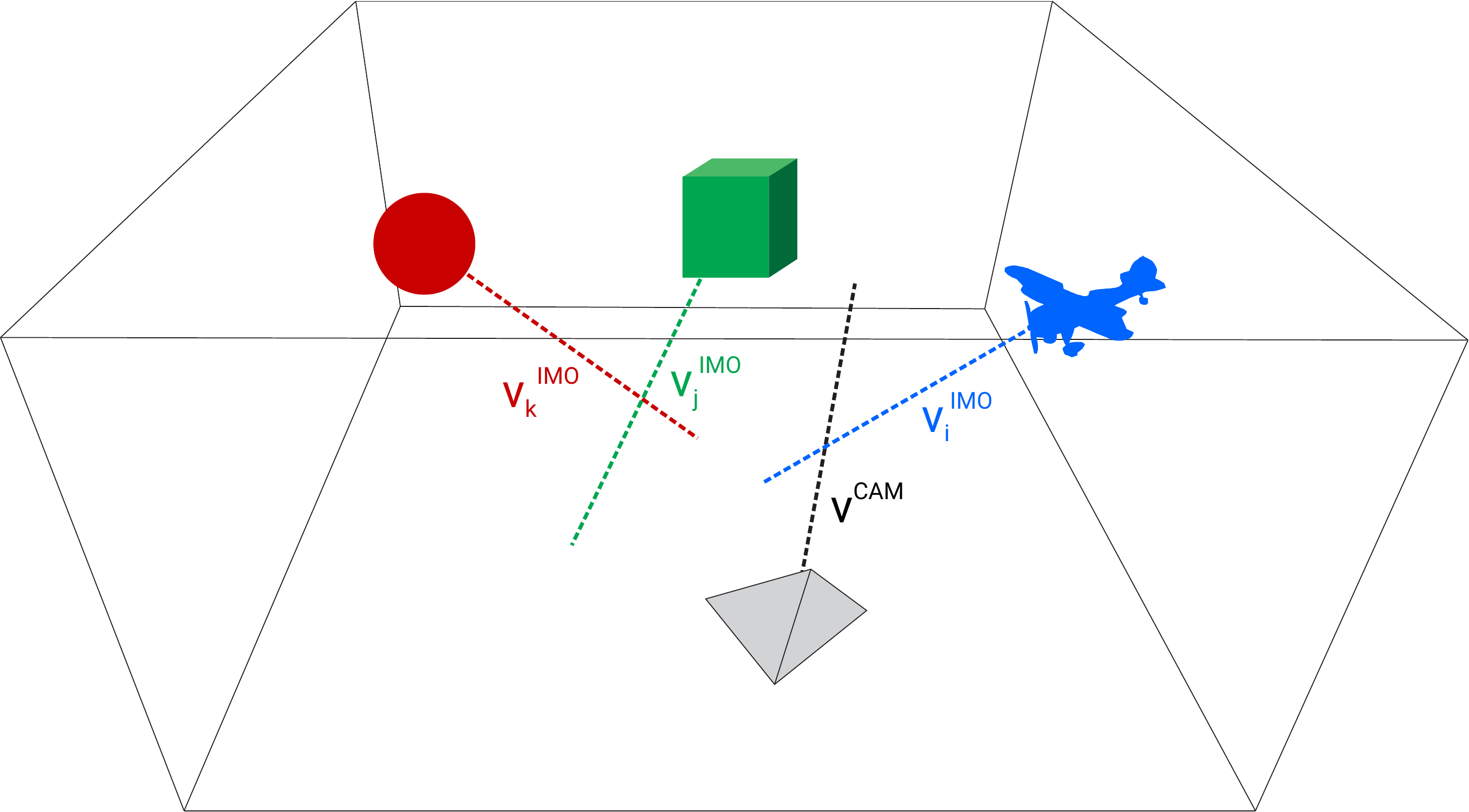}
\end{center}
% \vspace{-1.3\baselineskip}
   \caption{\small{Velocity vectors $v^\text{CAM}$ and $v^\text{IMO}_i$ used for data stratification in the MOD++ dataset.}}
\label{fig:MOD++}
\end{figure}

\section{MOD++ Dataset/Benchmark}
Currently, three main datasets exist for IMO segmentation using event cameras, namely, Extreme Event Dataset (EED)\cite{iROSBetterFlow} and EV-IMO\cite{mitrokhin2019ev} featuring real data and the synthetic Moving Object Dataset (MOD)\cite{sanket2019evdodgenet}. However none of the datasets have data stratification based on camera and/or object motion and/or velocities. To this end, we extend the MOD dataset presented in \cite{sanket2019evdodgenet} which we call MOD++ to add additional synthetic sequences (Refer to Table \ref{tab:DatasetTable}). This data stratification is explained next. Let the instantaneous velocity of the center of mass of the camera and IMOs be denoted as $v^\text{CAM}$ and $v^\text{IMO}_i$, where $i$ is the IMO index (Fig. \ref{fig:MOD++}). Now consider the angle and relative-magnitude between two vectors ($a, b$) denoted by $\theta$ and $\eta$ respectively and defined as follows: $\theta\left(a,b\right) = \cos^{-1}\left( \frac{a\cdot b}{\norm{a}\norm{b}}\right)$ and $\eta\left(a,b\right) = \frac{\norm{a}}{\norm{b}}$. Also, let the instantaneous rotational velocity around it's principal axes be denoted by $\omega$ where the superscripts and subscripts have the same meaning as that of linear velocities. We classify the sequences as follows:
1. Different linear velocities: Here we set the instantaneous rotational velocity close to zero, i.e.,  $\norm{\omega^\text{CAM}} \approx 0$ and $\norm{\omega^\text{IMO}_i} \approx 0$. We classify the motions based on relative angle and speed. If $\theta\left(v^\text{CAM},v^\text{IMO}_i\right) \in [0, 45]^\circ \forall i$ we call this sequence small angle. If $\theta\left(v^\text{CAM},v^\text{IMO}_i\right) \in [60, 120]^\circ \forall i$ we call this sequence medium angle and if $\theta\left(v^\text{CAM},v^\text{IMO}_i\right) \in [140, 180]^\circ \forall i$ we call this sequence large angle. (Note that we wrap the angles in the range [0, 180] in our case).

If $\eta\left(v^\text{IMO},v^\text{CAM}_i\right) \in [0.5, 3.0) \ \forall i$ we call this sequence slow. If $\eta\left(v^\text{IMO},v^\text{CAM}_i\right) \in [3.0, 7.0) \ \forall i$ we call this sequence medium and if $\eta\left(v^\text{IMO},v^\text{CAM}_i\right) \in [7.0, 10.0] \ \forall i$ we call this sequence fast.

If $\norm{\omega^\text{CAM} - \omega^\text{IMO}_i} \in [0, 5]^\circ s^{-1} \forall i$ we call this sequence slow rotation. If $\norm{\omega^\text{CAM} - \omega^\text{IMO}_i} \in [25, 30]^\circ s^{-1} \forall i$ we call this sequence medium rotation and if $\norm{\omega^\text{CAM} - \omega^\text{IMO}_i} \in [90, 100]^\circ s^{-1} \forall i$ we call this sequence fast rotation.

To make it easy to identify the sequence we use the following naming convention: \texttt{SeqSEQNUM\_ATTR1\_ATTRN} where \texttt{SEQNUM} is the sequence number which will determine the scene setup (room walls and objects with texture), \texttt{ATTR1} to \texttt{ATTRN} are modifiers which specify speed and/or rotation classifications. We use the following modifiers: \texttt{AS, AM, AL} for small, medium and large angles, \texttt{SS, SM, SL} for small, medium and large linear speeds, \texttt{RS, RM, RL} for small, medium and large rotational speeds.  For eg., \texttt{Seq4\_AM\_SL\_RS} would be the fourth sequence with medium angles, large linear speeds and small rotational speeds.

Additionally, we also provide two challenge sequences for researchers to evaluate their algorithm on: \texttt{Cube} and \texttt{Cup}. The \texttt{Cube} sequence is two cubes (a smaller cube on top of a larger cube) falling on the ground and breaking into smaller non-cube pieces. The \texttt{Cup} sequence is a bullet hitting a cup and shattering it into smaller fragments of different shapes.

% TODO: Show Fig. of various speeds with event images?
% TODO: Exploding and collision sequences
% TODO: Cluster membership is color-coded

\begin{table*}
    \centering
    \caption {Overview Of Related Datasets.}
     \label{tab:DatasetTable}
    \resizebox{\textwidth}{!}{

    \begin{tabular}{ccccc}
        \toprule
           & MOD++ & EV-IMO\cite{mitrokhin2019ev} & EED\cite{iROSBetterFlow} & MOD\cite{sanket2019evdodgenet}\\
          \hline
          Year & 2020 & 2019 & 2018 & 2019 \\
          \hline
          Data-type & Simulated & Real & Real & Simulated \\
          \hline
          Camera & Sim. DAVIS346C & DAVIS346C & DAVIS240B & Sim. DAVIS346C \\

          \hline
          Data & \makecell{Events \\ RGB Images @ 1000 Hz \\ 6-DoF Camera Poses @ 1000 Hz \\ 6-DoF IMO Poses @ 1000 Hz \\ IMO Bounding Boxes @ 1000 Hz \\ IMO Masks @ 1000 Hz \\ Optical Flow @ 1000 Hz \\ Depth @ 1000 Hz} & \makecell{Events \\ Grayscale Images @ 40 Hz \\ 6-DoF Camera Poses @ 200 Hz \\ 6-DoF IMO Poses @ 200 Hz \\ IMO Masks @ 40 Hz} & \makecell{Events \\ Grayscale Images @ 20 Hz \\  IMO Bounding Boxes @ 20 Hz} & \makecell{Events \\ RGB Images @ 1000 Hz \\ 6-DoF Camera Poses @ 1000 Hz \\ 6-DoF IMO Poses @ 1000 Hz \\ IMO Bounding Boxes @ 1000 Hz \\ IMO Masks @ 1000 Hz}\\
          \hline

          \makecell{Poses Ground \\ Truth (Acc.)} & Blender$^\text{\textregistered}$ Engine (Sub. mm) & \makecell{12 $\times$ Vicon$^\text{\textregistered}$ \\Vantage V8 Cameras ($\approx$ 1mm)} & -- & Blender$^\text{\textregistered}$ Engine (Sub. mm) \\

          \hline

          \makecell{IMO Bounding Boxes \\ (Masks) Ground Truth} & \makecell{Blender$^\text{\textregistered}$ Engine\\ for both} & \makecell{Scanned 3D Objects\\ projected using\\ Ground Truth Pose} & Hand-labelled (--) & \makecell{Blender$^\text{\textregistered}$ Engine\\ for both} \\

          \hline
          Num. Sequences & 43 & 30 & 5 & 7 \\
          \hline
          \makecell{Num. Unique Objects \\(Max. Number of Objects\\ in frame)} & 12 (9) & 4 (3) & 7 (3) & 9 (3) \\
          \hline
          Num. Backgrounds & 11 & 5 & 5 & 9 \\
          \hline
          Challenging Scenes & Exploding and Breaking objects & Fast camera motion & Extreme illumination & -- \\
          \hline
          Data Stratification & \makecell{Velocity Direction\\ Velocity Magnitude \\ IMO Rotation Magnitude} & -- & -- & -- \\
        \bottomrule
\end{tabular}}

\end{table*}

% \begin{figure*}[h!]
% \begin{center}
%     \includegraphics[width=\textwidth]{0-MMS/images/Merging.png}
% \end{center}
% \vspace{-1.3\baselineskip}
%   \caption{\small{Iterative model fitting and merging approach. The colors indicate the average temporal gradient for that particular cluster. (Blue indicates a low value and red indicates a high value). We stop merging whenever the average temporal gradient increases drastically after a merging step since this indicates a merger of the background with an IMO cluster.}}
% \label{fig:merge}
% \end{figure*}

% \subsubsection{Global model fitting}
% \subsubsection{Residual feature detection and tracking}
% \subsubsection{Initial clustering}
% \subsubsection{Sparse Model fitting}
% \subsubsection{Model merging based on spatio-temporal consistency}
% \subsubsection{Stopping criterion}

\section{Experiments and Results}

We evaluate our approach on publicly available real and synthetic datasets. We demonstrate our approach's performance both qualitatively and quantitatively employing two different metrics based on the availability of groundtruth information.

\begin{figure}[t!]
\begin{center}
    \includegraphics[width=\columnwidth]{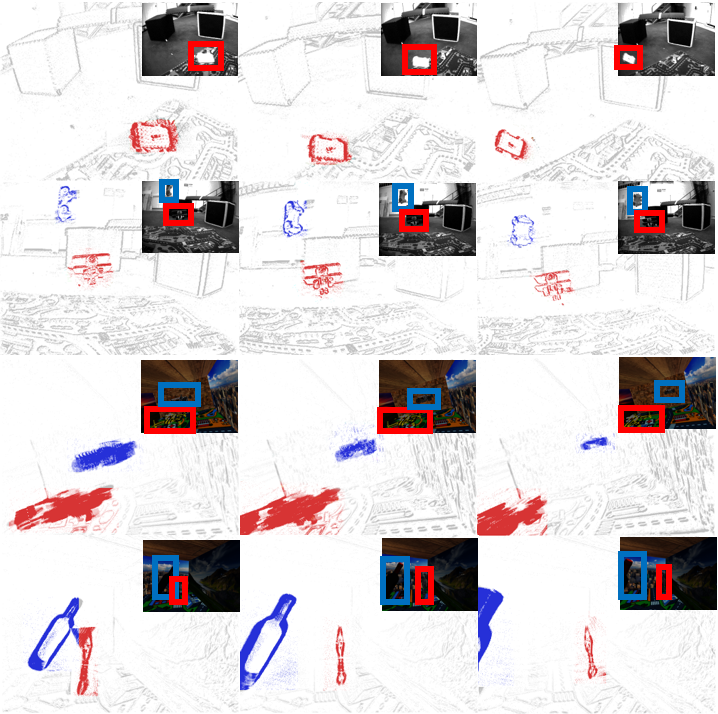}
\end{center}
% \vspace{-1.3\baselineskip}
   \caption{\small{Qualitative Evaluation of our method on three datasets. Top two rows: EV-IMO dataset, Bottom two rows: MOD dataset. Insets show the corresponding grayscale/RGB images for reference. The cluster membership is color coded where gray color indicates background cluster.  Bounding boxes on the images are color coded with respect to the objects for reference.}} % All the outputs are generated with the same value of $\lambda$.}
\label{fig:dataset}
\end{figure}

\subsection{Detection Rate}
For datasets which provide timestamped bounding boxes for the objects, we consider the prediction as success when the estimated bounding box fulfills two conditions; (1) it has a overlap of more than atleast 50\% with the groundtruth bounding box, (2) the area of intersection with the groundtruth box is higher than the intersection with outside area. We can formulate the metric as:
\begin{equation}
    \text{Success}\,\,\,\,\text{if}\,\,\,\,\mathcal{D}\cap \mathcal{G} > 0.5 \quad \text{and} \quad (\mathcal{D} \cap \mathcal{G}) > (\lnot\mathcal{G} \cap \mathcal{D})
    \label{overlap_equation}
\end{equation}
where $\mathcal{D}$ is the predicted mask and $\mathcal{G}$ is the groundtruth mask. We evaluate our pipeline's performance on all the datasets using this metric. We obtain the bounding box for our method by obtaining the convex hull on the cluster of events. For comparison purpose we evaluate the  performance of \cite{iROSBetterFlow} using the same metric on all the three datasets. For datasets with more than one sequence, we compute the average of each model's performance on individual sequences.% Since \cite{stoffregen2019event} does not have  publicly available code we could not evaluate the performance for the  MOD and EV-IMO dataset.
\begin{figure*}[t!]
    \centering
    \includegraphics[width=0.97\textwidth]{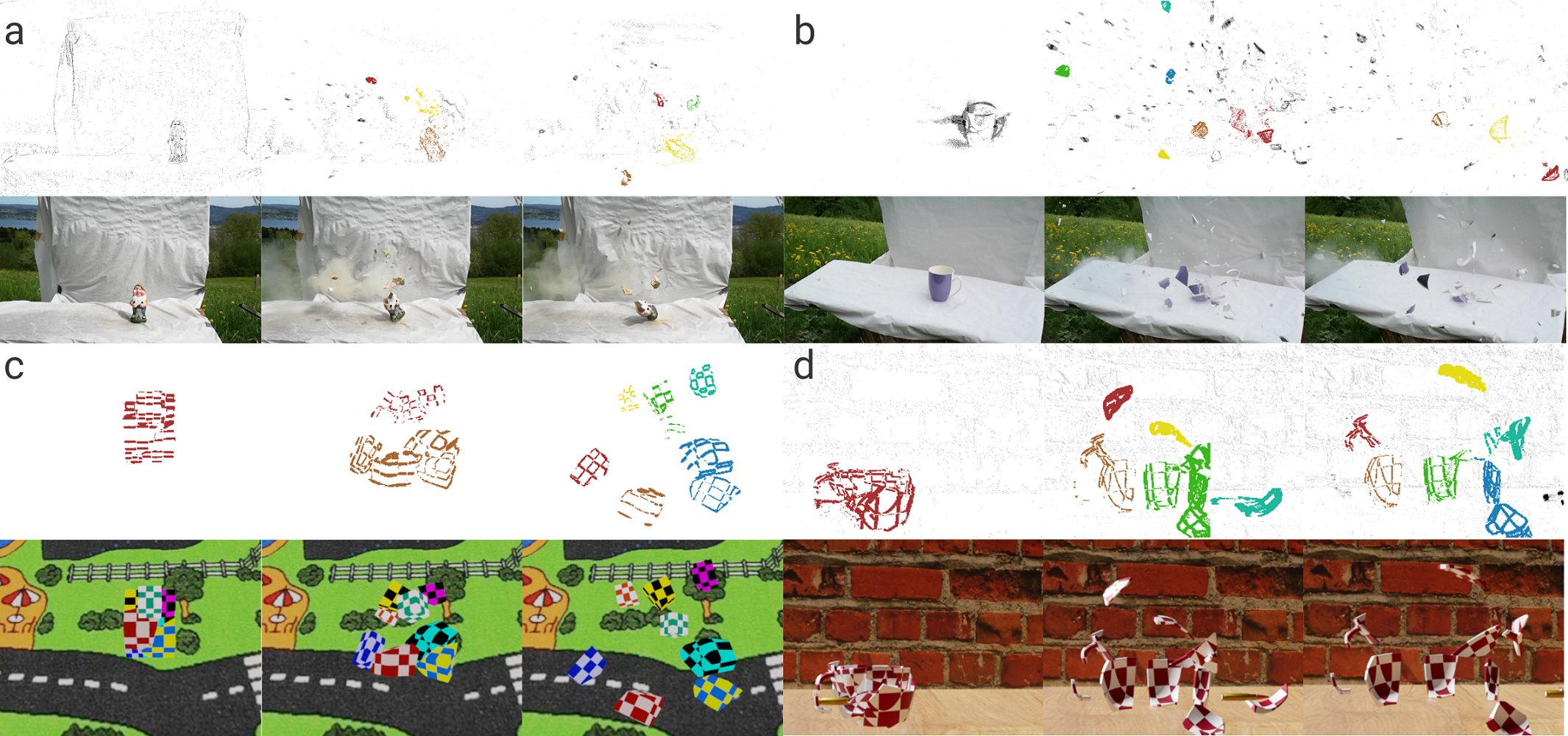}
    \caption{Multi-Motion Segmentation on the real sequences from \cite{rebecq2019high}. (a) \texttt{Gnome shooting}: Gnome statue getting shot by a bullet, (b) \texttt{Mug shooting}: Mug getting shot by a bullet.
    Challenge sequences from the proposed MOD++ dataset. (c) \texttt{Cube} breaking into smaller pieces by falling, and (d) \texttt{Cup}   getting shot by a bullet. The event frames are colored by cluster membership with gray showing background cluster. Note that the corresponding RGB frames \textit{are not used for computation and are provided for visualization purposes only}.}
    \label{fig:MOMSWithEventsBannerRealAndSim}
\end{figure*}

% TODO: Add rows
\begin{table*}[ht]
    \centering
     \caption {Comparison with state-of-the-art using the detection rate for different sequences of MOD++.}
     \label{tab:MOD++}
    \resizebox{\textwidth}{!}{
    \begin{tabular}{lllllllllll}
    \hline
         \multirow{2}{*}{Method} & \multicolumn{8}{c}{Detection Rate (DR in \%) $\uparrow$} & Speed $\uparrow$ & \multirow{2}{*}{Speed$\times$Avg. DR $\uparrow$}\\
         & \texttt{AS\_SM\_RS} & \texttt{AM\_SM\_RS} & \texttt{AL\_SM\_RS} & \texttt{AM\_SS\_RS} & \texttt{AM\_SL\_RS} & \texttt{AL\_SS\_RS} & \texttt{AL\_SS\_RM} & \texttt{AL\_SS\_RL} & (MEv/s) &  \\
         \hline
         Mitrokhin \etal\cite{iROSBetterFlow} & 35.24 & 32.29 & 38.12 & 28.78 & 43.28 & 24.56 & 32.29 & 39.65 & 5.41 & 185.47 \\
         EVDodgeNet \cite{sanket2019evdodgenet} & 42.25 & 46.94 & 53.23 & 37.81 & 61.72 & 46.13 & 43.50 & 52.38 & \underline{10.01} & \underline{480.42}\\

         $k$-Means (k=5) & 44.89 & 47.73 & 49.35 & 40.17 & 59.52 & 42.19 & 45.71 &  55.83 & 1.07 & 51.54\\

         $k$-Means (k=10) & \underline{60.36} & \underline{64.87} & \underline{59.27} & \underline{47.73} & \underline{65.78} & \underline{48.92} & \underline{54.74} & \underline{58.47} & 1.02 & 58.66\\
         $k$-Means (k=20) & 56.1 & 62.28 & 58.01 & 45.25 & 61.25 & 44.71 & 49.37 & 54.91 & 0.98 & 52.90 \\
         \hline
         Ours (No Propagation) & \textbf{70.13} & 73.29 & \textbf{72.58} & \textbf{65.80} & \textbf{85.76} & \textbf{66.24} & \textbf{72.90} & \textbf{79.02} & 0.83 & 60.77 \\
         Ours & 69.52 & \textbf{73.94} & 71.27 & 63.58 & 84.21 & 64.93 & 71.18 & 78.37 & \textbf{1.16} & \textbf{83.67}\\
         \hline
    \end{tabular}}
\end{table*}

\subsection{Intersection Over Union (IoU)}
IoU is one the most common and henceforth the most standard measure to evaluate and compare the performance of different segmentation methods. % We threshold the IoU to 0.5 for inference purposes.
IoU is given by:
\begin{equation*}
    IoU = \left(\mathcal{D} \cap \mathcal{G}\right)/\left(\mathcal{D} \cup \mathcal{G}\right)
    \label{IoU_equation}
\end{equation*}
% where $\mathcal{D}$ is the predicted mask and $\mathcal{G}$ is the groundtruth mask.
Our method outputs a cluster of events which are associated with an object. For the purpose of comparison we convert the sparse mask to a dense mask by assigning all the points lying inside the cluster as the same value. % Among all the datasets available, we believe that EV-IMO dataset is the most challenging. Hence, we consider to compare our other approaches with our approaches on EV-IMO. % As discussed earlier for EV-IMO, we evaluate our model only on the box and wall sequence. Since \cite{mitrokhin2019ev} code is not publicly available and their model trained on these two sequences, we do not consider their performance for comparison purposes here.

\subsection{Discussion of Results}

%TODO - Runtime comparison with Stoffregen
%TODO - Define object wrt IOU

Table \ref{tab:MOD++} reports results on our proposed MOD++ dataset. We pick eight scenarios with different relative (camera and IMO) velocity direction, velocity magnitude and rotation magnitude. We illustrate the merits of split and merge, and motion propagation through extensive ablation studies and compare with previous approaches. Our approach outperforms others by at least $\sim$10\% . Executing split and merge at every step offers better accuracy than propagating motion models (evident for more challenging scenarios like slow relative motion, large/small relative velocity direction and/or small rotation magnitude). However, motion propagation offers a speed-up without a significant loss of performance ($\sim$1\%). Even though simple thresholding \cite{iROSBetterFlow} and the deep learning-based approach \cite{sanket2019evdodgenet} offer better speed-up and Speed$\times$Avg. DR (a metric proposed in \cite{sanket2020prgflow}), their accuracies are almost 2-3$\times$ lower than our approach and is not reliable in challenging scenarios. We leave the speeding-up of our approach using deep learning to enable deployment on mobile robots for future work.

%TODO -

Table \ref{tab:DetectionRateResults} reports the result of our method in comparison with two state-of-the-art IMO detection methods \cite{iROSBetterFlow}, \cite{stoffregen2019event} using only a monocular event camera. Our method outperforms the previous methods by up to $\sim$12\% detection rate. Specifically, we outperform \cite{iROSBetterFlow} by a large margin (up to $\sim$32\%) on all the three datasets. % Also, we perform comparably with \cite{stoffregen2019event} on the EED dataset, even without knowing the number of moving objects, which proves the robustness of our approach in merging clusters.
% As our approach takes inspiration from Hierarchical clustering and we fit the motion model to only a subset of events (around the features) as opposed to every event in the cluster \cite{stoffregen2019event}. Also, contrast and distance functions are employed to merge clusters (a group of events) rather than an update of event-cluster assignments. Hence, our approach is faster than event-wise EM segmentation approach \cite{stoffregen2019event}.
Our approach is about 2$\times$ faster than \cite{stoffregen2019event} because of motion propagation while maintaining similar/slightly better accuracy on EED.

\begin{table}
    \centering
    \caption {Comparison with state-of-the-art using detection rate for EED, MOD, EV-IMO datasets.}
    \label{tab:DetectionRateResults}
    \begin{tabular}{ccccc}
    \toprule
    \multirow{2}{*}{Method} & \multicolumn{3}{c}{Detection rate for dataset (\%) $\uparrow$} & Speed $\uparrow$\\
         & EED & MOD & EV-IMO & (MEv/s)\\
         \hline
         Mitrokhin \etal \cite{iROSBetterFlow}  & 88.93 & 70.12 & 48.79 & 5.41 \\
          Stoffregen \etal \cite{stoffregen2019event} &  93.17 & - & - & 0.64$^*$ ($N_l$=10)\\
          Ours & \textbf{94.2} & \textbf{82.35} & \textbf{81.06} & 1.16 \\
         \bottomrule
    \end{tabular}
    \tiny{$^*$Results taken directly from \cite{stoffregen2019event}}
\end{table}
%TODO: Do we need to mention that stoffregen paper doesnt have the opensource code

Table \ref{tab:IoUResults} reports the comparison with two deep learning methods for IMO segmentation  \cite{mitrokhin2019ev} and \cite{sanket2019evdodgenet} using the IoU metric. % Note that, both approaches are deep learning based.
\cite{sanket2019evdodgenet} was trained on the MOD dataset and is tested here on the EV-IMO dataset without any fine-tuning/re-training. We outperform \cite{sanket2019evdodgenet} and \cite{mitrokhin2019ev} (which was trained on EV-IMO) on the EV-IMO dataset. % And,  was trained on the EV-IMO dataset and our approach improves the IoU by a small margin.

\begin{table}
    \centering
    \caption{Comparison with state-of-the-art using IoU for EV-IMO.}
    \label{tab:IoUResults}
    \begin{tabular}{ccc}
    \toprule
         Method & IoU\\
         \hline
         EV-IMO~\cite{mitrokhin2019ev}  & 77.00$^*$\\
         EVDodgeNet~\cite{sanket2019evdodgenet} & 65.76 \\
         Ours & \textbf{80.37}\\
         \bottomrule
    \end{tabular}\\
    \tiny{$^*$Results taken directly from \cite{mitrokhin2019ev} in which \texttt{boxes} and \texttt{wall} are used for training.}
\end{table}

Fig. \ref{fig:dataset} shows qualitative results of our approach on the two datasets (top two rows show results for the EV-IMO dataset and the last two rows show results for the MOD dataset). Gray areas in the event images show the background cluster and red/blue colored regions show the differently segmented IMOs. The outputs show the robustness of our approach to shape, size and speed of the objects and in-variance with respect to camera motion. Also, note that the objects are sometimes very hard to detect in the corresponding grayscale/RGB frames in Fig. \ref{fig:dataset} motivating the use of event cameras for IMO detection using motion cues.

\begin{figure}[t!]
\begin{center}
    \includegraphics[width=1.0\columnwidth]{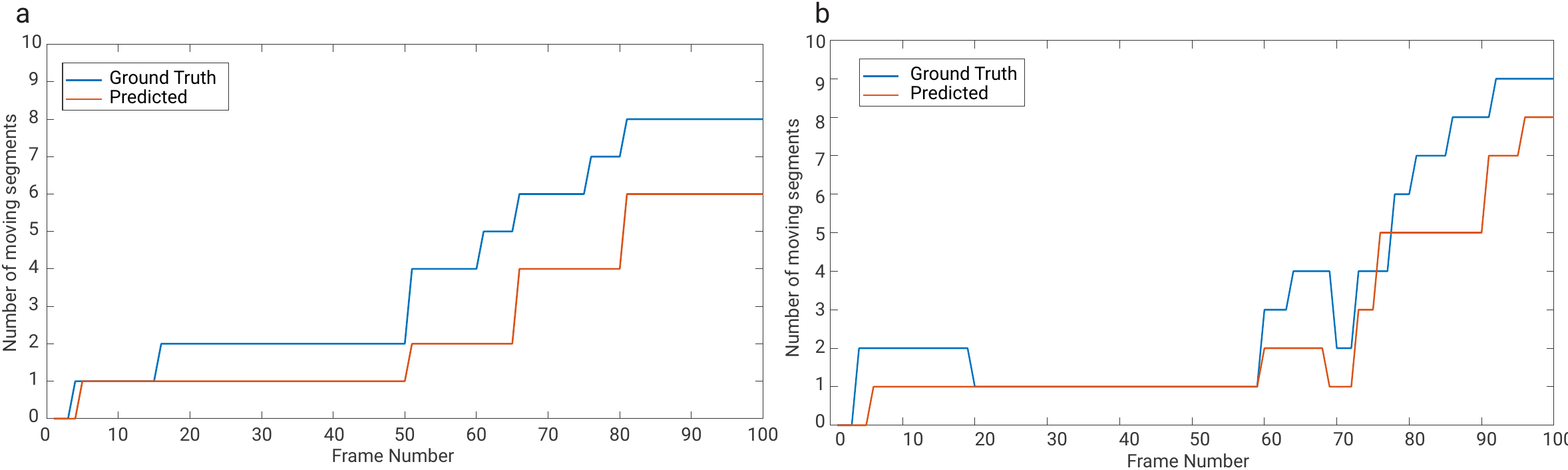}
\end{center}
% \vspace{-1.3\baselineskip}
   \caption{Number of moving segments vs. frame number (time) on challenge sequences of MOD++: (a) \texttt{Cube}, (b) \texttt{Cup}.}
\label{fig:cup_cube}
\end{figure}

We obtain the ground truth IMO by counting ground truth labels with IoU overlap $\leq$0.2. The graph shows robustness of our approach with increasing number of moving segments. The predicted number of segments closely matches the ground truth. Segmenting solely based on motion with a monocular event-camera is ambiguous in challenging scenarios and results could be improved with incorporation of depth and appearance information in our split and merge step which we believe is the logical next step for future work.

Figs. \ref{fig:MOMSWithEventsBannerRealAndSim}{\color{red}a} and \ref{fig:MOMSWithEventsBannerRealAndSim}{\color{red}b} show the output of our method for challenging sequences of \texttt{Gnome shooting} and \texttt{Mug shooting} from \cite{rebecq2019high} showing that our method performs well even on real sequences with a large number of objects.

Fig. \ref{fig:cup_cube} shows performance of our algorithm with the respect to number of moving segments across time on challenge sequences of MOD++ dataset i.e., \texttt{Cube} and \texttt{Cup} sequence (shown in Figs. \ref{fig:MOMSWithEventsBannerRealAndSim}{\color{red}c} and \ref{fig:MOMSWithEventsBannerRealAndSim}{\color{red}d}).

 % Our approach takes $\sim$0.4 ms for the model fitting, $\sim$13 ms for feature extraction and tracking, $\sim$5 ms for splitting and merging, and $\sim$5 ms for motion propagation on a single thread i7 CPU and NVIDIA Titan Xp.
 Our algorithm runs on a hybrid CPU and GPU system (i7 CPU and NVIDIA Titan Xp GPU). Model fitting and feature extractions are run on GPU in parallel. Even though our core algorithm runs fast, the bottleneck is in the memory transfer to and from the GPU. \textit{The complexity of our approach is linear in the number of clusters, events and frequency of cluster keyslices initiation.} Table \ref{tab:MOD++} shows the speed of our algorithm in comparison with other approaches in Million Events per second (MEv/s). Our motion propagation and keyslicing provides a speed-up of upto 40\% without compromising accuracy.

% Our approach takes $\sim$20 ms for the model fitting, $\sim$13 ms for feature extraction and tracking, $\sim$25 ms for splitting and merging, and $\sim$5 ms for motion propagation on a single thread i7 CPU and NVIDIA Titan Xp. Most of heavy of computations such as model fitting and feature extractions are run on GPU for speed-up.

\section{Conclusions and Future Work}
We presented a method for multi-motion segmentation using data from a monocular event camera. Our approach works by splitting the scene into smaller motions and then iteratively merging them based on a contrast measure. To our knowledge, this is the first approach for monocular independent motion segmentation which combines a bottom-up feature tracking and top-down motion compensation into a unified pipeline. We further speed up our method by using the concept of motion propagation and cluster keyslices.

A comprehensive qualitative and quantitative evaluation is provided on three challenging event motion segmentation datasets, namely, EV-IMO, EED and MOD showcasing the robustness of our approach. Our method outperforms the previous state-of-the-art approaches by upto $\sim$12\% detection, thereby achieving the new state-of-the-art on the three aforementioned datasets. To accelerate further research in this area, we present and open-source a new benchmark dataset MOD++ which includes challenging scenes such as cube breaking and a mug getting shot by a bullet along with extensive data stratification in-terms of camera and object motion, velocity magnitudes, direction and rotational speeds. We achieve 73.21\% detection rate on MOD++ which is 2 to 3$\times$ higher than the state-of-the-art methods.

\bibliographystyle{unsrt}
\bibliography{references}

\begin{thebibliography}{10}

\bibitem{fermuller1993navigational}
Cornelia Ferm\"uiiller.
\newblock Navigational preliminaries.
\newblock {\em Active Perception}, 1:103--150, 1993.

\bibitem{sanket2018gapflyt}
Nitin~J Sanket, Chahat~Deep Singh, Kanishka Ganguly, Cornelia Ferm{\"u}ller,
  and Yiannis Aloimonos.
\newblock Gapflyt: Active vision based minimalist structure-less gap detection
  for quadrotor flight.
\newblock volume~3, pages 2799--2806. IEEE, 2018.

\bibitem{liang2019salientdso}
Huai-Jen Liang, Nitin~J Sanket, Cornelia Ferm{\"u}ller, and Yiannis Aloimonos.
\newblock Salientdso: Bringing attention to direct sparse odometry.
\newblock {\em IEEE Transactions on Automation Science and Engineering},
  16(4):1619--1626, 2019.

\bibitem{lichtsteiner2008128}
Patrick Lichtsteiner, Christoph Posch, and Tobi Delbruck.
\newblock A 128$\times$ 128 120 db 15$\mu$s latency asynchronous temporal
  contrast vision sensor.
\newblock {\em IEEE journal of solid-state circuits}, 43(2):566--576, 2008.

\bibitem{iROSBetterFlow}
Anton Mitrokhin, Cornelia Ferm{\"u}ller, Chethan Parameshwara, and Yiannis
  Aloimonos.
\newblock Event-based moving object detection and tracking.
\newblock {\em IEEE/RSJ Int. Conf. Intelligent Robots and Systems (IROS)},
  2018.

\bibitem{litzenberger2006}
M.~{Litzenberger}, C.~{Posch}, D.~{Bauer}, A.~N. {Belbachir}, P.~{Schon},
  B.~{Kohn}, and H.~{Garn}.
\newblock Embedded vision system for real-time object tracking using an
  asynchronous transient vision sensor.
\newblock In {\em IEEE 12th Digital Signal Processing Workshop and 4th IEEE
  Signal Processing Education Workshop}, pages 173--178, 2006.

\bibitem{linares2015usb3}
Alejandro Linares-Barranco, Francisco G{\'o}mez-Rodr{\'\i}guez, Vicente
  Villanueva, Luca Longinotti, and Tobi Delbr{\"u}ck.
\newblock A usb3. 0 fpga event-based filtering and tracking framework for
  dynamic vision sensors.
\newblock In {\em 2015 IEEE International Symposium on Circuits and Systems
  (ISCAS)}, pages 2417--2420. IEEE, 2015.

\bibitem{mishra2017saccade}
Abhishek Mishra, Rohan Ghosh, Jose~C Principe, Nitish~V Thakor, and Sunil~L
  Kukreja.
\newblock A saccade based framework for real-time motion segmentation using
  event based vision sensors.
\newblock {\em Frontiers in neuroscience}, 11:83, 2017.

\bibitem{barranco2018real}
Francisco Barranco, Cornelia Ferm\"uller, and Eduardo Ros.
\newblock Real-time clustering and multi-target tracking using event-based
  sensors.
\newblock In {\em 2018 IEEE/RSJ International Conference on Intelligent Robots
  and Systems (IROS)}, pages 5764--5769, 2018.

\bibitem{gallego2017accurate}
Guillermo Gallego and Davide Scaramuzza.
\newblock Accurate angular velocity estimation with an event camera.
\newblock {\em IEEE Robotics and Automation Letters}, 2(2):632--639, 2017.

\bibitem{zhu2017event}
A.~Z. {Zhu}, N.~{Atanasov}, and K.~{Daniilidis}.
\newblock Event-based visual inertial odometry.
\newblock In {\em 2017 IEEE Conference on Computer Vision and Pattern
  Recognition (CVPR)}, pages 5816--5824, July 2017.

\bibitem{gallego2019focus}
Guillermo Gallego, Mathias Gehrig, and Davide Scaramuzza.
\newblock Focus is all you need: Loss functions for event-based vision.
\newblock In {\em Proceedings of the IEEE Conference on Computer Vision and
  Pattern Recognition}, pages 12280--12289, 2019.

\bibitem{Stoffregen_2019_CVPR}
Timo Stoffregen and Lindsay Kleeman.
\newblock Event cameras, contrast maximization and reward functions: An
  analysis.
\newblock In {\em The IEEE Conference on Computer Vision and Pattern
  Recognition (CVPR)}, June 2019.

\bibitem{stoffregen2019event}
Timo Stoffregen, Guillermo Gallego, Tom Drummond, Lindsay Kleeman, and Davide
  Scaramuzza.
\newblock Event-based motion segmentation by motion compensation.
\newblock In {\em Proceedings of the IEEE International Conference on Computer
  Vision}, pages 7244--7253, 2019.

\bibitem{barranco2015bio}
Francisco Barranco, Cornelia Ferm{\"u}ller, and Yiannis Aloimonos.
\newblock Bio-inspired motion estimation with event-driven sensors.
\newblock In {\em International Work-Conference on Artificial Neural Networks},
  pages 309--321. Springer, 2015.

\bibitem{sanket2019evdodgenet}
Nitin~J. Sanket, Chethan~M. Parameshwara, Chahat~Deep Singh, Ashwin~V.
  Kuruttukulam, Cornelia Fermüller, Davide Scaramuzza, and Yiannis Aloimonos.
\newblock Evdodgenet: Deep dynamic obstacle dodging with event cameras, 2019.

\bibitem{mitrokhin2019ev}
Anton Mitrokhin, Chengxi Ye, Cornelia Ferm\"uller, Yiannis Aloimonos, and Tobi
  Delbruck.
\newblock {EV-IMO:} motion segmentation dataset and learning pipeline for event
  cameras.
\newblock In {\em IEEE/RSJ Int. Conf. Intelligent Robots and Systems (IROS)},
  2019.

\bibitem{mitrokhin2020learning}
Anton Mitrokhin, Zhiyuan Hua, Cornelia Fermuller, and Yiannis Aloimonos.
\newblock Learning visual motion segmentation using event surfaces.
\newblock In {\em Proceedings of the IEEE/CVF Conference on Computer Vision and
  Pattern Recognition}, pages 14414--14423, 2020.

\bibitem{Gallego2018AUC}
Guillermo Gallego, Henri Rebecq, and Davide Scaramuzza.
\newblock A unifying contrast maximization framework for event cameras, with
  applications to motion, depth, and optical flow estimation.
\newblock {\em IEEE Conf. Computer Vision and Pattern Recognition (CVPR)},
  2018.

\bibitem{seok2020robust}
Hochang Seok and Jongwoo Lim.
\newblock Robust feature tracking in dvs event stream using b{\'e}zier mapping.
\newblock In {\em The IEEE Winter Conference on Applications of Computer
  Vision}, pages 1658--1667, 2020.

\bibitem{gehrig2020eklt}
Daniel Gehrig, Henri Rebecq, Guillermo Gallego, and Davide Scaramuzza.
\newblock Eklt: Asynchronous photometric feature tracking using events and
  frames.
\newblock {\em International Journal of Computer Vision}, 128(3):601--618,
  2020.

\bibitem{dardelet2018event}
Laurent Dardelet, Sio-Hoi Ieng, and Ryad Benosman.
\newblock Event-based features selection and tracking from intertwined
  estimation of velocity and generative contours.
\newblock {\em arXiv preprint arXiv:1811.07839}, 2018.

\bibitem{alzugaray2019asynchronous}
Ignacio Alzugaray and Margarita Chli.
\newblock Asynchronous multi-hypothesis tracking of features with event
  cameras.
\newblock In {\em 2019 International Conference on 3D Vision (3DV)}, pages
  269--278. IEEE, 2019.

\bibitem{detone18superpoint}
Daniel DeTone, Tomasz Malisiewicz, and Andrew Rabinovich.
\newblock Superpoint: Self-supervised interest point detection and description.
\newblock In {\em CVPR Deep Learning for Visual SLAM Workshop}, 2018.

\bibitem{rebecq2019high}
Henri Rebecq, Ren{\'e} Ranftl, Vladlen Koltun, and Davide Scaramuzza.
\newblock High speed and high dynamic range video with an event camera.
\newblock {\em IEEE Transactions on Pattern Analysis and Machine Intelligence},
  2019.

\bibitem{sanket2020prgflow}
Nitin~J Sanket, Chahat~Deep Singh, Cornelia Ferm{\"u}ller, and Yiannis
  Aloimonos.
\newblock {PRGFlow: Benchmarking SWAP-Aware Unified Deep Visual Inertial
  Odometry}.
\newblock {\em arXiv preprint arXiv:2006.06753}, 2020.

\end{thebibliography}

\end{document}